\documentclass[lettersize,journal]{IEEEtran}
\usepackage{cite}
\usepackage{amsmath}
\usepackage{amssymb}
\usepackage{algorithmic}
\usepackage{array}
\usepackage[pdftex]{graphicx}
\usepackage{caption}
\usepackage[most]{tcolorbox}
\usepackage{color,soul}
\soulregister\cite7
\soulregister\ref7
\soulregister\tablename7
\soulregister\pageref7
\usepackage{adjustbox}
\DeclareGraphicsExtensions{.pdf}
\usepackage{subcaption}
\usepackage{multirow}
\usepackage{balance}
\usepackage{booktabs}
\usepackage[switch]{lineno} 

\ifCLASSOPTIONcompsoc
\usepackage[caption=false,font=normalsize,labelfont=sf,textfont=sf]{subfig}

\else
 \usepackage[caption=false,font=footnotesize]{subfig}

\usepackage{stfloats}
\usepackage{url}
\begin{document}


\title{RouteFormer: A Transformer-Based Routing Framework for Autonomous Vehicles}

\author{
    Yazan Youssef, \IEEEmembership{Member, IEEE}, Paulo Ricardo Marques de Araujo, \IEEEmembership{Member, IEEE}, Aboelmagd Noureldin, \IEEEmembership{Senior Member, IEEE}, and Sidney Givigi, \IEEEmembership{Senior Member, IEEE}%
    \thanks{Y. Youssef is with the Department
    of Electrical and Computer Engineering, Queen's University, Kingston, ON, Canada (e-mail: yazan.youssef@queensu.ca).}%
    \thanks{A. Noureldin is with the Department of Electrical and Computer Engineering at the Royal Military College of Canada, and the Department of Electrical and Computer Engineering at Queen's University, both in Kingston, Canada (e-mail: aboelmagd.noureldin@rmc.ca).}%
    \thanks{S. Givigi and P. Araujo are with the School of Computing, Queen's University, Kingston, ON, Canada (e-mail: \{sidney.givigi, paulo.araujo\}@queensu.ca).}%
}

\maketitle

\begin{abstract}
Autonomous surveillance missions in Internet of Things (IoT) networks often involve solving NP-hard combinatorial optimization problems to ensure efficient resource utilization. To address the limitations of conventional heuristics in dynamic environments, we propose RouteFormer, a novel framework for single-agent routing in graph-based terrains. RouteFormer creates a synergy between the global context awareness of the transformer self-attention mechanism and the adaptive decision-making capabilities of Reinforcement Learning (RL). This architecture allows the system to output optimized routing decisions that adapt to complex task dependencies and resource availability without requiring labeled training datasets. We evaluated RouteFormer on varying graph sizes designed to resemble realistic reconnaissance missions. The results indicate that our model effectively handles the complexity of missions requiring multiple action profiles, outperforming baseline approaches, in terms of both time and distance. Specifically, RouteFormer achieved 10\% and 7\% reduction in distance compared to the solutions obtained from well-established solvers like Concorde and  Lin-Kernighan-Helsgaun-3 (LKH-3). This improvement was achieved by effectively incorporating mission-specific constraints that traditional solvers overlook. The proposed framework serves as a modular, scalable pipeline for diverse autonomous scheduling and routing tasks.

\end{abstract}


\IEEEpeerreviewmaketitle

\section{Introduction}
\IEEEPARstart{E}{fficient} navigation and autonomous path planning are fundamental challenges in Internet of Things (IoT) domains, particularly for mobile agents deployed in surveillance, reconnaissance, and submarine missions~\cite{IoT_Cooperative_PathPlanning, IoT_ReconnaissanceScheduling, IoT_ocean_PathPlanning}. Optimizing the trajectories of these autonomous agents is essential for meeting operational requirements, such as minimizing travel time and adhering to strictly limited energy budgets~\cite{IoT_UAV_PathPlanning}.

Traditional approaches model these missions as complex combinatorial optimization problems, often variants of the Traveling Salesperson Problem (TSP) or Vehicle Routing Problem (VRP), requiring sophisticated algorithms to determine the optimal sequence of locations to visit~\cite {2024_heauristics_weakness, IoT_JointDeployment, IoT_2019_TruckAssisted_routing}.

Over the past decade, routing strategies have advanced significantly, particularly through the use of greedy algorithms and constructive heuristics~\cite{2020_annual_reviews}. However, while computationally inexpensive, these methods often converge to sub-optimal local minima and struggle to adapt to the dynamic constraints typical of modern IoT environments~\cite{2024_heauristics_weakness}.

In the search for higher quality solutions, researchers have utilized exact optimization techniques like integer linear programming~\cite{ILP} and constraint programming~\cite{CP}, as well as meta-heuristic algorithms such as genetic algorithms~\cite{GA} and simulated annealing~\cite{SA}. While these methods have shown promise for static TSP instances, they suffer from poor scalability and prohibitive computational latency when applied to large-scale, real-time routing problems~\cite{2017_Neural_Combinatorial}. These limitations have prompted the adoption of machine learning approaches, specifically Neural Combinatorial Optimization (NCO) via reinforcement learning (RL), which learns to generate routing policies directly from experience~\cite{2017_Neural_Combinatorial}.

Concurrently, the emergence of the transformer architecture has revolutionized fields like natural language processing and sequential data modeling~\cite{2020_NLP,bubeck2023sparks}. Originally developed for machine translation~\cite{attention_2017s}, transformers serve as the backbone of state-of-the-art models like BERT and GPT~\cite{bubeck2023sparks}. Their unique self-attention mechanisms enable the modeling of long range dependencies, making them exceptionally well suited for routing problems, which can be framed as sequential decision-making tasks where the ``sentence'' is the path and the ``words'' are the visited nodes~\cite{TSP_RL_2022}.

Inspired by these advancements, researchers are increasingly exploring transformers for spatial representation and routing optimization~\cite{2022_TA_RL}. The transformer's ability to capture global graph structures and complex node relationships, combined with the adaptability of RL, offers a powerful alternative to traditional solvers~\cite{2024_RL_AGV_TA}.

The primary contribution of this paper is RouteFormer, a novel routing framework that leverages transformer architecture to optimize multi-stage mission profiles for single agent missions in graph-based environments. Trained end-to-end using reinforcement learning, this framework is, to the best of our knowledge, the first of its kind to address the specific complexities of multi-stage, time-extended routing problems where node costs are dynamic. Unlike traditional combinatorial solvers that degrade in speed as complexity increases, RouteFormer generates high quality solutions with inference times in the order of seconds, regardless of the problem configuration. To demonstrate its efficacy and generalization capabilities, we apply RouteFormer to a Unmanned Aerial Vehicle (UAV)-based Search and Rescue (SAR) scenario. In this context, the system successfully optimizes flight paths across a series of target locations with varying spatial extents.

The paper is organized as follows. Section~\ref{sec:RelatedWork} reviews machine learning approaches for routing and path planning, while identifying gaps on a systematic approach for the application of such approaches. Section~\ref{sec:ProblemFormulation} presents the routing problem mathematically. Section~\ref{sec:RouteFormer} details RouteFormer. Section~\ref{sec:SimulationSetup} describes the simulation setup and the tested scenarios. Section~\ref{sec:Results} presents the simulation results; and Section~\ref{sec:Conclusion} concludes the paper.

\section{Related Work}\label{sec:RelatedWork}
\subsection{Neural Combinatorial Optimization for Routing}
Traditional exact solvers (e.g., Concorde) and meta-heuristics (e.g., Genetic Algorithms) have long been the standard for routing problems like the TSP. However, they struggle with computational scalability and dynamic constraints in real-time environments~\cite{shahbazian2024integrating}. Recently, NCO has emerged as a paradigm to approximate NP-hard routing problems using deep learning~\cite{NCO}. Early works utilized Recurrent Neural Networks and Pointer Networks to output node sequences~\cite{2017_Neural_Combinatorial}. However, the introduction of the Transformer architecture has revolutionized this field. Kool et al.~\cite{2019_kool_attention} introduced the Attention Model, establishing that self-attention mechanisms are naturally suited for graph-based routing problems.

More recent advancements have focused on improving the generalization of these models to complex, heterogeneous environments. For instance, Li et al.~\cite{Li_2024_MultiType} introduced a multi-type attention mechanism that adapts the transformer architecture to handle the distinct structural constraints of multi-depot routing, demonstrating superior transfer capabilities over static baselines. In the context of resource-constrained missions, Wang et al.~\cite{Wang_2024_EVRP} extended NCO to the Electric Vehicle Routing Problem, 
employing a Graph Attention Network 
to dynamically manage battery levels -- a critical spatiotemporal constraint analogous to the time limits in IoT surveillance. Unlike these works, which often focus on specific vehicle physics or logistical benchmarks, our framework adapts transformer capabilities specifically for mission-oriented constraints, which involve multi-stage decision making.

\subsection{Task Assignment and Scheduling in IoT}
Distinct from routing, task assignment focuses on the high-level allocation of resources to tasks without necessarily optimizing the exact traversal path. In IoT networks, this is often framed as optimizing resource utility or minimizing latency. Recent research has heavily leveraged Deep Reinforcement Learning (DRL) for these high-level decisions. For instance, Xu et al.~\cite{Xu_2024_TransEdge} proposed TransEdge, a framework utilizing Graph Neural Networks (GNNs) and DRL to optimize task offloading decisions in edge-computing transportation systems, successfully minimizing response latency in dynamic environments. In the specific domain of UAVs, Lyu et al.~\cite{Lyu_2024_UAV} introduced an RL-assisted framework for multi-UAV task allocation in Industrial IoT, which dynamically assigns inspection tasks to agents while managing conflict-free path constraints.

\subsection{Routing and Planning for Autonomous Agents}
Beyond general theoretical models, significant research efforts have been dedicated to application-specific scheduling and routing within multi-vehicle and IoT-enabled missions. For instance, Niu et al.\cite{Niu_2022} employed a Deep Q-Network to simulate a control center for UAV task scheduling in disaster scenarios. In the context of data collection, Hu et al.\cite{IoT_joint_routing} proposed a framework utilizing an Adaptive Memory Procedure for path planning with vehicle-assisted drones. Addressing heterogeneous constraints, Bai et al.\cite{2023_package_delivery_TA_drone} applied heuristic algorithms to solve capacity-constrained delivery problems. Given the prevalence of drones in IoT logistics, Deng et al.\cite{IoT_2024_UAV_delivery} developed a Public Transportation System assisted algorithm for multi-drone routing. Similarly, Wu et al.\cite{IoT_Cooperative_PathPlanning} combined Estimation of Distribution Algorithms with Genetic Algorithms to optimize path planning for IoT surveillance tasks. Recognizing the limitations of static heuristics, recent studies have increasingly integrated Reinforcement Learning (RL) to handle dynamic environments. For example, Xi et al.\cite{IoT_ocean_PathPlanning} utilized RL alongside comprehensive oceanographic data for Autonomous Underwater Vehicle planning, while Li et al.~\cite{2024_RL_AGV_TA} employed an RL-trained network to address warehouse order collection by framing it as a Multiple Depot TSP.

The discussed works deal with routing problems where the nodes to visit are static points with fixed service costs. However, in realistic IoT missions like surveillance and reconnaissance, a ``task'' is rarely a simple waypoint; rather, it often represents multiple decisions for covering a spatially extended area, such as a varying search zone, where the agent must make internal decisions regarding coverage and execution. The cost to complete such a task is not fixed but depends on factors like entry angles and scanning complexity, creating a multi-stage optimization challenge. Standard TSP-based or heuristic solvers fail to capture this granularity, often decoupling the global routing from the local task execution logic. In contrast, our framework reformulates the problem as a holistic mission optimization. By leveraging the transformer architecture, we allow the agent to jointly reason about the global sequence of visits and the specific internal complexities of each task area. This enables RouteFormer to minimize the total mission cost, integrating both global traversal time and local execution overheads, more effectively.

\section{Problem Formulation} \label{sec:ProblemFormulation}
Motion planning for robotic missions is typically hierarchically decomposed into path planning, trajectory planning, and path tracking~\cite{path_planning_2023}. \textit{Path planning} addresses the spatial constraints of the mission, generating a collision-free geometric route from a start to a goal configuration. \textit{Trajectory planning} subsequently assigns a time law to this geometric path, governing velocity and acceleration profiles. Path tracking then functions as the low-level feedback controller, ensuring the robot’s physical state converges with the generated reference. This work specifically addresses the \textit{path planning} domain.

The problem we solve with RouteFormer can be formulated similarly to the TRP. Given a graph $G=(V,E)$, $V=\{1,2,\cdots,n\}$ is a set of areas to be visited, where $\{0\}$ is the arbitrary initial point of the path, and $E=\{(i,j):i,j\in V, i\neq j\}$ is the connections between areas. The objective is for the agent to visit all the areas once with the shortest path. Note that here we use ``shortest path'' in a loose sense, as this does not refer necessarily to distance, but to a cost, which can be time, energy, or another metric. For simplicity and compliance to the literature, from now on we use ``distance'' to mean ``cost''. In addition, agent does not necessarily refer to a UAV, it refers to the robot used for the specified mission, as kinematics are out of the scope of this paper.

Since the areas are not dimensionless, i.e., we are not dealing with only points, the distance $d_{ij}$ between two different areas $i$ and $j$, with $(i,j) \in E$ can be different depending on the starting and end points chosen by the agent. Therefore, the distance is defined as $d_{ij} = \|s_j - e_i\|$, i.e., the Euclidean distance between the selected starting point in area $j$ and the endpoint in area $i$.

The selection of start and end points for the trajectory is crucial for minimizing distance. The area can have any shape and the agent can enter or leave the area anywhere depending on the chosen or predefined movement pattern for that area. A scenario might consist of multiple areas with different properties requiring specific scanning methods. For example, dense tree canopies may need a compact zig-zag pattern to maximize coverage, whereas open areas could benefit from sparser movement patterns like spirals or stars. Another scenario involves the flexibility to select any movement pattern for any area, and this is the focus of this work.

Therefore, given $G$ and any set of possible movement patterns, the scheduling system must determine the shortest path such that the agent (1) visits each area exactly once, (2) starts and ends at the same area, and (3) accounts for the movement pattern chosen for each area.

\section{RouteFormer}\label{sec:RouteFormer}
RouteFormer addresses a single-agent scheduling problem with multiple decisions, as described in Section~\ref{sec:ProblemFormulation}. Using a map of the region of interest as input, RouteFormer employs artificial neural networks to generate an optimized task sequence, minimizing execution costs, such as distance and time.
This approach is similar to the TRP, where minimizing distance is equivalent to minimizing time if the agent travels at a constant speed, with the pattern selection affecting the service time for each area. The pattern selection determines the landing point of the agent, which in turn affects the distance and time that the agent needs to reach the next area.

\begin{figure}
    \begin{center}
        \includegraphics[width=\columnwidth]{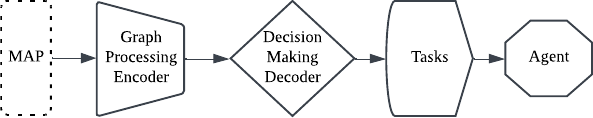}
    \end{center}
    \caption{System workflow.}
    \label{fig:framework}
\end{figure}

Based on Joshi et al.'s pipeline~\cite{TSP_RL_2022}, RouteFormer adapts key components to meet the specific objectives of this problem. The system overview is illustrated in \figurename~\ref{fig:framework}, with its components detailed in the following sections.

\subsection{Input Map}\label{subsec: Map} 
\begin{figure}
    \centering
     \includegraphics[width=\columnwidth]{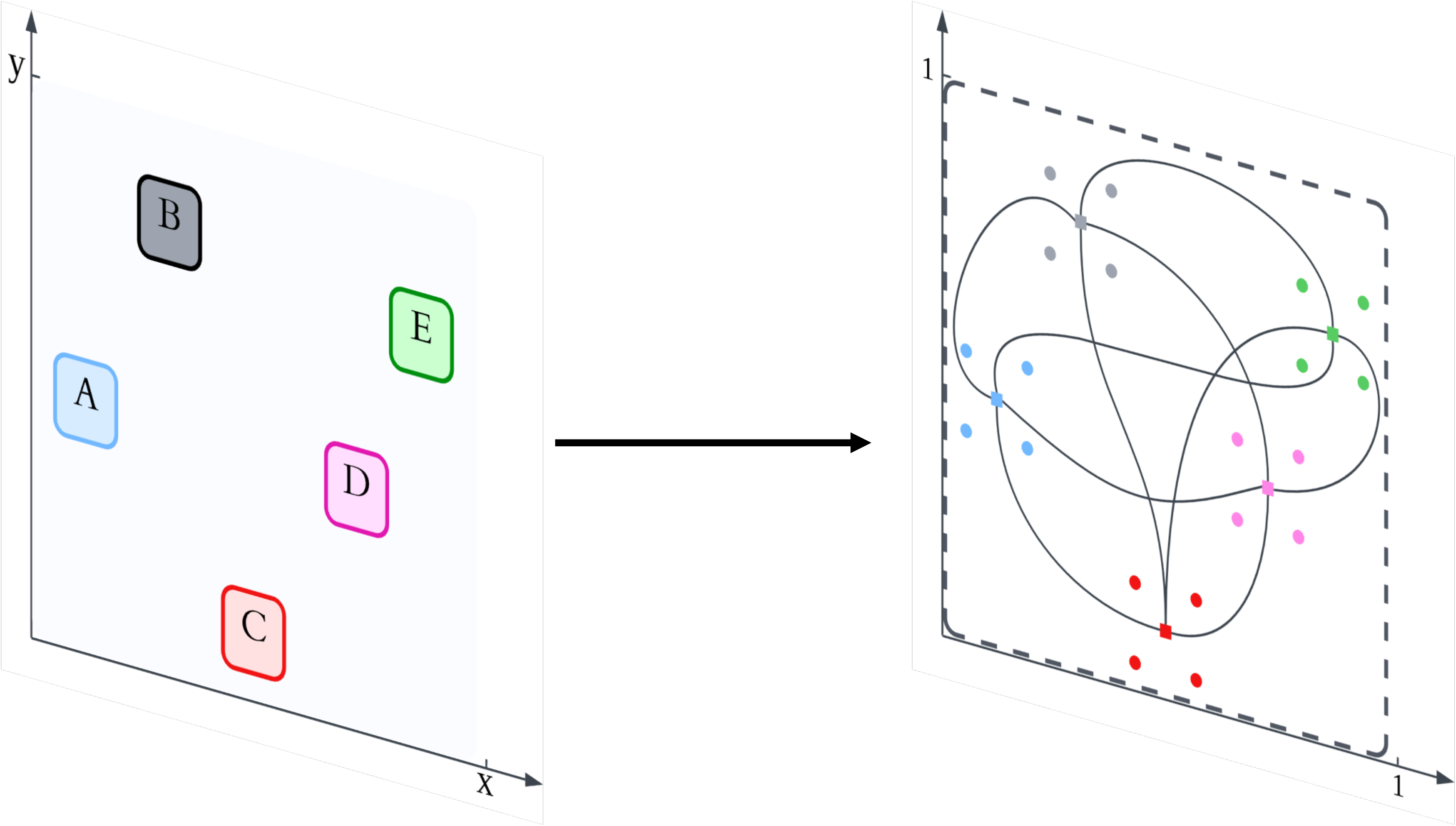}
     \caption{Input: The map is normalized and the required features are extracted}
     \label{fig:input}
\end{figure}

A map indicating the areas that need to be scanned is inputted into the system. Geometric approximations for the areas (e.g., rectangular shapes) are used to build the input map, capturing and describing the locations and features of the areas that must be visited.
Each input map is composed of $3$ matrices: adjacency matrix $A \in \mathbb{R}^{n\times n}$, feature matrix $X \in \mathbb{R}^{n\times f}$, and a position matrix $P \in \mathbb{R}^{n\times 2}$, where $n$ and $f$ are the number of areas and the number of features, respectively.
For example, in the demonstration scenario detailed in section~\ref{sec:scenario}, the areas were defined by rectangular frames using $4$ corners determining their borders. Therefore, $f$ was set to $8$, corresponding to four pairs of ($x,y$) coordinates, as shown in \figurename~\ref{fig:input}.

The adjacency matrix $A$ describes the connectivity of the areas, and the feature matrix $X$ contains the features used for describing the areas. The feature matrix is used to calculate $d_{ij}$. Finally, the position matrix $P$ contains the centers of the areas in the map, which are the same as the centers of the frames.

\subsection{Encoder} \label{sec:Encoder}
\begin{figure*}
     \centering
     \includegraphics[width=\linewidth]{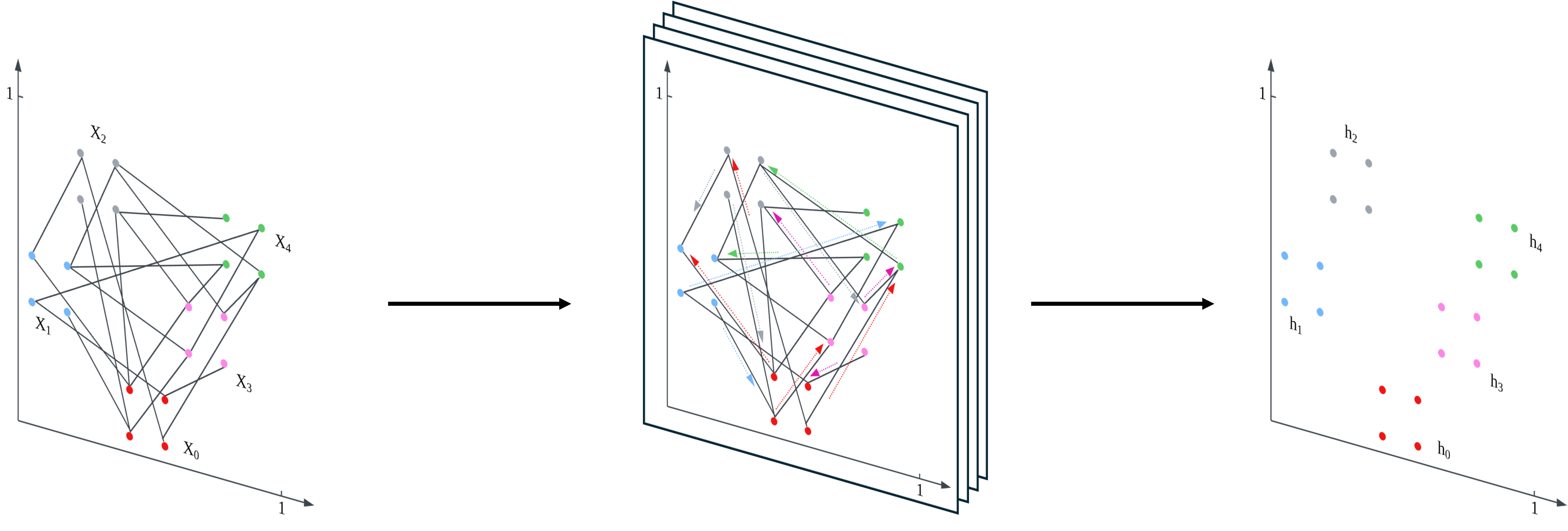}
     \caption{Encoder: The encoder processes the input features through GNN to obtain a new feature representation}
     \label{fig:encoder}
\end{figure*}

The encoder, in \figurename~\ref{fig:encoder}, transforms the input map into a $d_1$-dimensional representation using a GNN. Specifically, it converts the feature matrix $X \in \mathbb{R}^{n \times f}$ into $X_1 \in \mathbb{R}^{n \times d_1}$ with a Gated Graph ConvNet~\cite{GConvNET}. This GNN, comprising $L$ layers, uses anisotropic aggregation and a gating mechanism to refine node and edge features through recursive message passing, improving graph representation.Hence, the node feature $h_{i}^\ell$ and edge feature $e_{ij}^\ell$ at layer $\ell$ are:
\begin{eqnarray}
\begin{aligned}
&h_i^{\ell+1} = h_i^{\ell} + \\
&\operatorname{ReLU}\Biggl(\operatorname{BN}\Biggl(U^{\ell} h_i^{\ell} + \operatorname{AG}_{j \in \mathcal{N}_i}\Biggl(\sigma\left(e_{i j}^{\ell}\right) \odot V^{\ell} h_j^{\ell}\Biggr)\Biggr)\Biggr),
\end{aligned}
\label{h_l}
\end{eqnarray}

\begin{eqnarray}
\begin{aligned}
e_{ij}^{\ell+1} = &e_{ij}^{\ell} + \\
&\operatorname{ReLU}\Biggl(\operatorname{BN}\Biggl(A^{\ell} e_{ij}^{\ell}+ B^{\ell} h_i^{\ell} +C^{\ell} h_j^{\ell}\Biggr)\Biggr),
\end{aligned}
\label{e_l}
\end{eqnarray}
\noindent where $U^\ell, V^\ell, A^\ell, B^\ell, C^\ell \in \mathbb{R}^{d_1 \times d_1}$ are the learnable parameters of the encoder, $\operatorname{ReLU}$ represents the Rectified Linear Unit function, $\operatorname{BN}$ represents the batch normalization layer, $\operatorname{AG}$ is the neighborhood aggregation function, $\sigma$ is the sigmoid function, and $\odot$ is the Hadamard product. For the first layer ($\ell=0$), the inputs $h_i^0$ and $e_{ij}^0$ are $d_1$-dimensional linear projections of the feature matrix $X$ and the Euclidean distance between the centers of the areas $\| p_i -p_j\|$, respectively.


\subsection{Decoder}\label{sec:decoder}
\begin{figure*}
     \centering
     \includegraphics[width=\linewidth]{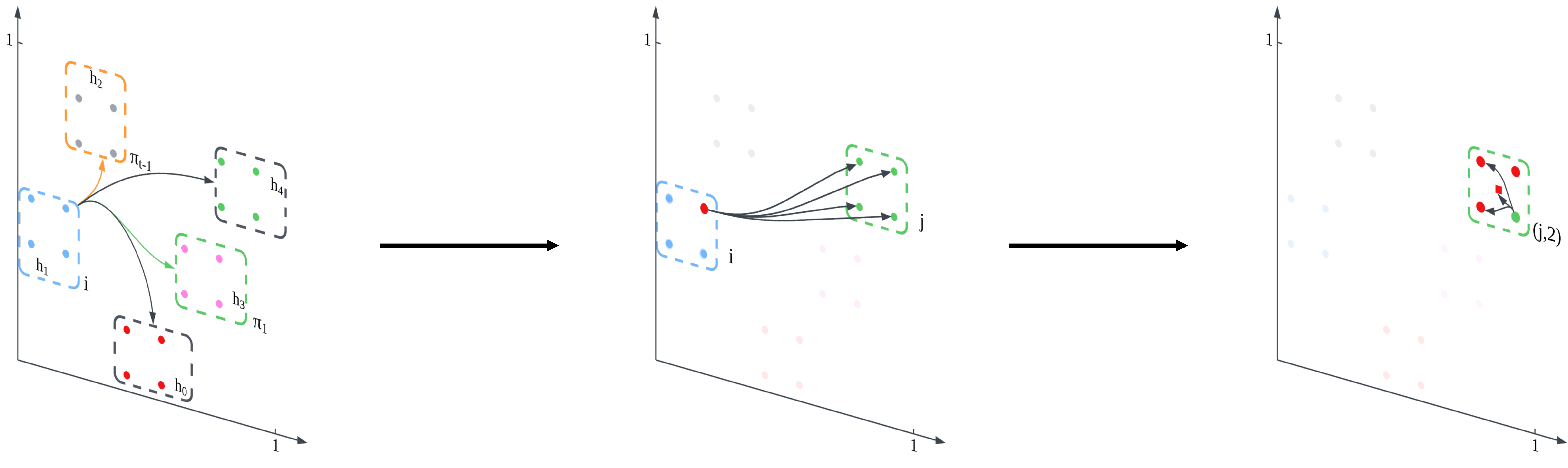}
     \caption{Decoder: The decoder processes the new features in 3 networks: the first network selects the next area to visit, the second network assigns the starting point in the chosen area, and finally the third network decides the movement pattern to be followed.}
     \label{fig:decoder}
\end{figure*}

The decision decoder, illustrated in \figurename~\ref{fig:framework}, employs an auto-regressive transformer architecture for decoding and decision-making. Transformers are effective for handling long sequences with dependencies \cite{bubeck2023sparks}. Therefore, the proposed decoder consists of three sequential attention networks (illustrated in \figurename~\ref{fig:decoder}): 
\begin{enumerate}
    \item \textbf{Network I} selects the next area to visit.
    \item \textbf{Network II} determines the starting point in the area.
    \item \textbf{Network III} decides the movement pattern for the area.
\end{enumerate}

Each one of these networks follows the sequence: Input Embedding, Multi-Head Attention, Embedding, Attention, and Softmax.

Network [I], which chooses the next area, uses the encoder's output to build an initial context representation $\hat{h}_i^C$ for the current node $i$ at time step $t$. This context, given by
\begin{equation}
\hat{h}_i^C = W_F \left[ h_G, h_{\pi_1^{\prime}}^L, h_{\pi_{t-1}^{\prime}}^L \right], \quad h_G = \frac{1}{n} \sum_{i=0}^{n} h_i^L,
\label{eq:h_c_init}
\end{equation}
\noindent includes learnable weights $W_F \in \mathbb{R}^{d_1 \times d_1}$ and averages the features from the encoder. Here, $h_{\pi_1^{\prime}}^L$ and $h_{\pi_{t-1}^{\prime}}^L$ are features of the first and last areas in the partial tour ($\pi^{\prime}$) up to $t-1$. This context $\hat{h}_i^C$ is used to highlight the correlation between the current area and the existing tour, as illustrated in the first component of \figurename~\ref{fig:decoder}.

The initial context \eqref{eq:h_c_init} is refined using Multi-Head Attention (MHA) over the node embeddings. We define the parameters of the first MHA ($\texttt{MHA}_1$) as $Q_1 = W_{C_1} \hat{h}_i^C$, $K_1 = \left\{W_{A_1}[h_1^L, \ldots, h_n^L]\right\}$, and $V_1 = \left\{W_{A_2}[h_1^L, \ldots, h_n^L]\right\}$, where $W_{C_1}, W_{A_1}, W_{A_2} \in \mathbb{R}^{d_1  \times d_1}$ are learnable parameters. $Q_1$ represents our current knowledge, while $K_1$ and $V_1$ represent the options for the output. The MHA operation is then given by:
\begin{equation}
h_i^C=\operatorname{MHA}_{1}(Q_1, K_1, V_1).
\label{eq:MHA_1}
\end{equation}

The unnormalized logit for choosing the next area $j$ (or edge $e_{ij}$) is computed using an attention mechanism between the context $h_i^C$ from \eqref{eq:MHA_1} and the embedding $h_j^L$:
\begin{eqnarray}
\hat{p}[A_j|A_i]= M \cdot \tanh \left(\frac{\left(W_{Q_1} h_i^C\right)^T \cdot\left(W_{K_1} h_j^L\right)}{\sqrt{d_1}}\right),
\label{eq:P_j}
\end{eqnarray}

\noindent where $j \neq \pi_{t}^{\prime}$. If $j$ is already part of the tour, $\hat{p}[A_j|A_i] = \infty$. $W_{Q_1}, W_{K_1} \in \mathbb{R}^{d_1 \times d_1}$ are learnable parameters, and $\tanh$ keeps the logits in the range $[-M, M]$. The logits are then converted to probabilities $p[A_j|A_i]$ via a softmax operation. 

This formulation, based on \cite{TSP_RL_2022}, addresses traditional TSPs. The following sections extend this to enable RouteFormer to perform task assignments and be readily extended to TRPs.

After selecting area $j$ using \eqref{eq:P_j}, the features of area $j$ are used to determine the starting point. As illustrated in the middle component of \figurename~\ref{fig:decoder}, the $(x,y)$ coordinates of the features of area $j$ and the current location coordinates are fed into the Network [II]. The current location is where the agent ended up in area $i$ at step $t-1$.

A context representation for the current location is created using MHA over the corner coordinates of area $j$. This context is used to correlate the agent's location to the possible starting locations in the chosen area. The coordinates of the current location and the features are embedded using a feedforward linear network, resulting in $d_2$-dimensional embeddings:
\[
h_{st} = W_{C_2}h_{current}, \quad W_{\Lambda}[h_1,h_2,h_3,h_4]
\]
\noindent where \( h_{current} = (x_{current}, y_{current}) \), \( [h_1, h_2, h_3, h_4] \) are the corner coordinates, and \( W_{C_2}, W_{\Lambda} \in \mathbb{R}^{d_2 \times 2} \) are trainable parameters. $\texttt{MHA}_2$ parameters are:
\begin{eqnarray*}
    Q_2 &=& h_{st},\\
    \quad K_2 &=& \left\{W_{\Lambda 1}[h_1,h_2,h_3,h_4]\right\}, \\
    \quad V_2 &=& \left\{W_{\Lambda 2}[h_1,h_2,h_3,h_4]\right\}.
\end{eqnarray*}

The MHA operation yields the context \( h_{st}^C \):
\begin{equation}
    h_{st}^C=\operatorname{MHA}_{2}(Q_2, K_2, V_2).
\label{eq: h_s_C}
\end{equation}

The unnormalized logit for choosing corner \( h_k \), \( k \in \{1,2,3,4\} \), as the starting point is calculated using an attention mechanism between \( h_{st}^C \) and \( h_k \):
\begin{equation}
\begin{aligned}
&\hat{p}[\Lambda_k|A_j]=
&M \cdot \tanh \left(\frac{\left(W_{Q_2} h_{st}^C\right)^T \cdot\left(W_{K_2} h_k\right)}{\sqrt{d_2}}\right),
\end{aligned}
\label{eq: P_k}
\end{equation}

Finally, the logits \( \hat{p}[\Lambda_k|A_j] \) are converted to probabilities \( p[\Lambda_k|A_j] \) using a softmax operation.

The final step of the decoder is to choose the pattern the agent should follow. This is done similarly to the previous steps. The coordinates of the stopping points for each pattern ($h_{P_z}$) along with the starting point $h_k$ obtained from Network [II], are fed into Network [III], as illustrated in the third component of \figurename~\ref{fig:decoder}.

First, a context representation for point $h_k$ is created using MHA over the possible stopping points associated with the patterns. Assuming there are $p$ possible patterns, the coordinates of $h_k$ and the stopping points from patterns $P_1, P_2, \ldots, P_p$ are embedded using a $d_3$-dimensional input embedding layer. $\texttt{MHA}_3$ parameters are:
\begin{eqnarray*}
    Q_3 &=& W_{C_3}[h_k], \\
    \quad K_3 &=& \left\{W_{\Psi_1}[h_{P_1},h_{P_2},\cdots,h_{P_p}]\right\},\\
    \quad V_3 &=& \left\{W_{\Psi_2}[h_{P_1},h_{P_2},\cdots,h_{P_p}]\right\}
\end{eqnarray*}
\noindent where \(W_{C_3}, W_{\Psi_1}, W_{\Psi_2} \in \mathbb{R}^{d_3 \times 2}\) are the trainable weight matrices. The context representation for selecting the pattern is then:
\begin{equation}
    h_k^C=\operatorname{MHA}_3(Q_3, K_3, V_3).
    \label{eq:hK}
\end{equation}

The unnormalized logit for selecting pattern $P_z$, $z \in \{1,2,\dots,p\}$, is computed using a final attention mechanism between the context $h_k^C$ (from \eqref{eq:hK}) and the coordinates of the stopping point from pattern $P_z$ ($h_{P_z}$):
\begin{equation}
\begin{aligned}
&\hat{p}[\Psi_z| A_j,\Lambda_k]=
&M \cdot \tanh \left(\frac{\left(W_{Q_3} h_k^C\right)^T \cdot\left(W_{K_3} h_{P_z}\right)}{\sqrt{d_3}}\right),
\end{aligned}
\label{eq:P_p}
\end{equation}
\noindent where $W_{Q_3} \in \mathbb{R}^{d_3 \times d_3}$ and $W_{K_3} \in \mathbb{R}^{d_3 \times 2}$ are learnable parameters. The logits $\hat{p}[\Psi_z| A_j, \Lambda_k]$ are converted to probabilities using softmax over all patterns. A pattern $\Psi_z$ is selected, and its associated stopping point becomes the current location for the next step. The final probability of the action (choosing area $A_j$, starting point $h_k$, and pattern $P_z$) is:
\begin{equation}
p[j,k,z]= p[A_j] \cdot p[\Lambda_k|A_j] \cdot p[\Psi_z| A_j, \Lambda_k].
\label{eq:P_a}
\end{equation}

This process repeats until all areas are covered. Consequently, the final output of the decoder is the sequence of tasks that will be followed by the agent.

\subsection{RL Policy Training}
RL is highly effective for training networks in scenarios where ground truth solutions are unavailable, as in the case of search and rescue task scheduling. In this paper, we use the REINFORCE policy gradient algorithm due to its simplicity. The objective is to minimize the total distance covered in a given scenario, which in turn would minimize the total execution time of the mission. The loss for an instance $s$ parameterized by $\theta$ is defined as $\mathcal{L}(\theta|s)=\mathbb{E}_{p_{\theta}(\pi|s)}[L(\pi)]$, where $L(\pi)$ is the tour length and $p_{\theta}(\pi|s)$ is the probability distribution of the tour $\pi$.

The gradient for minimizing $\mathcal{L}$ using REINFORCE is:
\begin{equation}
    \nabla \mathcal{L}(\theta|s) = \mathbb{E}_{p_{\theta}(\pi|s)}[(L(\pi)-b(s))\nabla \operatorname{log}(p_{\theta}(\pi|s))],
    \label{gradient}
\end{equation}
where $b(s)$ is a baseline to reduce gradient variance. Here, $L(\pi)$ is the tour length per epoch, and $p_{\theta}(\pi|s)$ is the probability of the tour, calculated as the product of action probabilities $p[j,k,z]$ from \eqref{eq:P_a}.
\section{Simulation Setup}\label{sec:SimulationSetup}
\subsection{Scenario} \label{sec:scenario}
To demonstrate RouteFormer's effectiveness, we tested the proposed system in a scenario that can illustrate a search-and-rescue application, where an agent has to visit several areas in the shortest distance/time possible. 
Without loss of generality, we represent the areas as squares, where they are defined by their four corners, resulting in a feature matrix $X \in \mathbb{R}^{n\times 8}$, corresponding to four pairs of ($x,y$) coordinates.

\begin{figure}
     \centering
     \begin{subfigure}[b]{0.3\columnwidth}
         \centering
         \includegraphics[width=\textwidth]{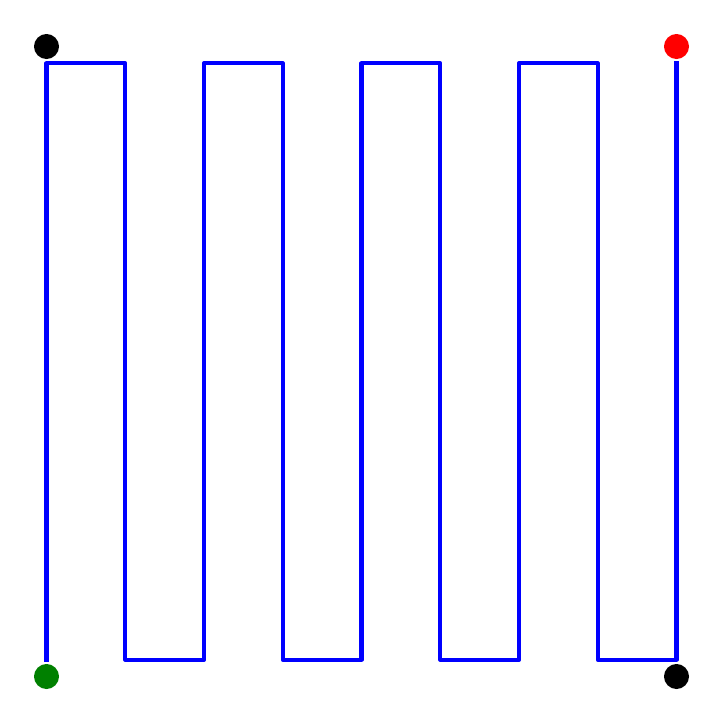}
         \caption{}
         \label{zv}
     \end{subfigure}
     \hfill
     \begin{subfigure}[b]{0.3\columnwidth}
         \centering
         \includegraphics[width=\textwidth]{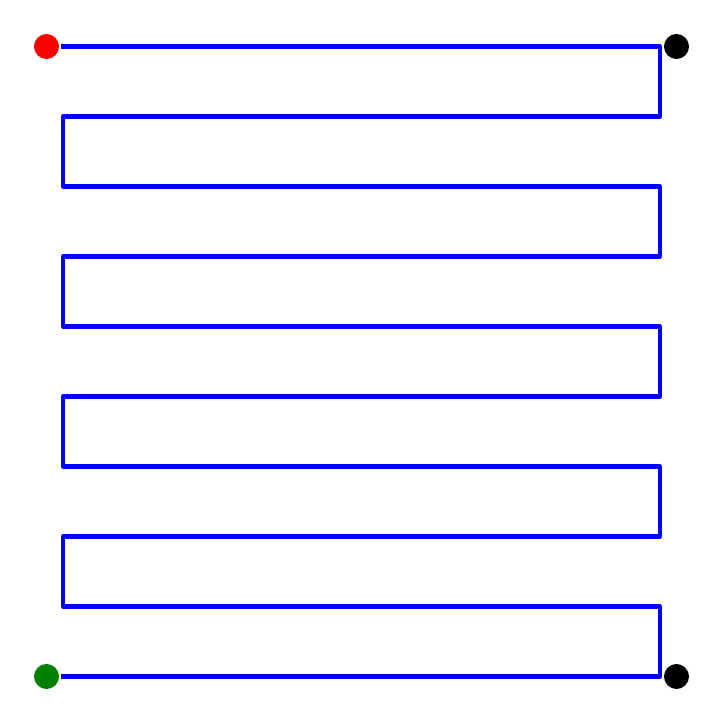}
         \caption{}
         \label{zh}
     \end{subfigure}
     \hfill
     \begin{subfigure}[b]{0.3\columnwidth}
         \centering
         \includegraphics[width=\textwidth]{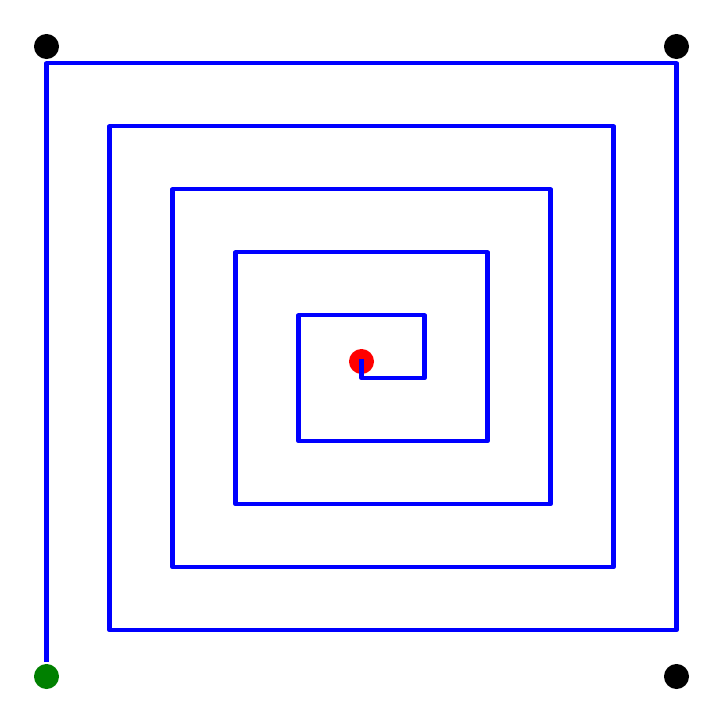}
         \caption{}
         \label{s}
     \end{subfigure}
\caption{Available patterns. (a) Vertical Zig-Zag, (b) Horizontal Zig-Zag, and (c) Spiral.}
\label{available_patterns}
\end{figure}

Additionally, we define three possible movement patterns for the agent: Vertical lawn mower, horizontal lawn mower, or spiral, as shown in \figurename~\ref{available_patterns}, where the green and red dots represent the starting and ending positions, respectively.

\subsection{Dataset}
For training, \( n \) areas, represented by frames, were generated within a unit square. Each area center was sampled from a uniform distribution $C = (c_x,c_y)$ where  $c_x, c_y \in [0,1]$. Then a random radius $r \in [0.01, 0.03]$ was added and subtracted from the center coordinates to form the four corners. This process was repeated until all $n$ areas were generated, ensuring they stayed within the unit square and did not overlap. The unit square was chosen for easy scalability. In total, the dataset comprises $128,000$ randomly generated maps. 

\subsection{Hyper-parameters}
For the encoder, $3$ layers were used, with hidden dimension $d_1=128$. Following the encoder, the decoder consists of the $3$ parts as discussed in Section~ \ref{sec:decoder}. For $P[A_j|A_i]$ given in \eqref{eq:P_j}, we used $d_1=128$ for the hidden dimension and $8$ heads for the MHA network. For $P[\Lambda_k|A_j]$ and $P[\Psi_z| A_j, \Lambda_k]$ given in \eqref{eq: P_k} and \eqref{eq:P_p}, respectively, we used $d_2=d_3=128$ for the hidden dimension and a single head for their respective MHA networks. The described configuration resulted in approximately $360,000$ trainable parameters.    

\subsection{Training Process}
The training was conducted for $100$ epochs, with each epoch generating $128,000$ maps, divided into $1,000$ batches of $128$. The model was trained using the Adam optimizer with a fixed learning rate of $10^{-4}$.
The RL environment was developed using GYM \cite{GYM}, chosen for its compatibility with the Robot Operating System (ROS) \cite{2017_gym_gazebo_ros, 2019_gym_gazebo}, facilitating easy future deployment on real robots.
The proposed model was implemented using PyTorch 2.1, an NVIDIA GeForce 3090, and an Intel i9 11900KF Eight-Core Processor 3.5GHz (5.3GHz With Turbo Boost).

\subsection{Evaluation Metrics and Compared Methods}
To evaluate the performance of the proposed method, the following tests were conducted:
\begin{enumerate}
    \item \textbf{TSP Validation:} The environment's behavior was verified using the dataset from \cite{TSP_RL_2022}, comparing results against the baseline. This test was used to verify the functionality of Network I.
    \item \textbf{Decoder Functionality:} Separate tests were conducted for the start/end points and pattern elements using two evaluation sets with 1,280 map samples of 15 and 20 areas. The entry and exit points were fixed for all the areas in the evaluation set. An edge weight matrix was then provided accordingly to the Concorde solver (optimal TSP solver known so far) and its results were used as baseline for comparison. Network III was disabled, by fixing the pattern, to verify the functionality of Network II when connected to Network I.
    \item \textbf{System Assessment:} The full system was finally tested against Concorde solver results. For this test, RouteFormer was expected to outperform Concorde by considering the cost of selecting new start/end points for patterns.
    \item \textbf{Scaling capabilities:} A final test with 30, 40 and 50 areas were conducted to evaluate the performance of the proposed model in previously unseen environments, i.e., more areas than the number of areas used to train the model.
\end{enumerate}

The evaluation metric used, known as the optimality gap, $o_g \in \mathbb{R}$, follows the standard approach in the literature~\cite{TSP_RL_2022} and is defined as
\begin{equation}
    o_g = [ 1 - \left(\frac{l_m}{l_s}\right)]\cdot 100,
\end{equation}
\noindent where $l_m, l_s \in \mathbb{R}$ are the model and solver tour lengths, respectively.

The solver used in these tests is the Concorde solver \cite{concorde}. Concorde's TSP solver is an LP-based program designed to solve TSP-like problems. It has been used to obtain the optimal solutions to the full set of 110 TSPLIB instances. Hence, its solution was used as the baseline solution when evaluating RouteFormer's performance, as it is known to provide \textit{optimal} solutions to TSP and TSP-like problems.

For the baseline in~\eqref{gradient}, we used a greedy rollout approach \cite{2019_kool_attention}. The model in step $t$ of the training is frozen and serves as the baseline updated every $u = 1000$ steps. Each epoch consists of 1000 steps, with the baseline updated at the end of each epoch. For real scenario testing (as discussed in item 4 above), we used a critic baseline \cite{2017_Neural_Combinatorial} for comparison. In the critic baseline, $b(s)$ in \eqref{gradient} is defined as $b(s) = \hat{v}(s,\textbf{w})$, where $\hat{v}(s,\textbf{w})$ is a learned value function parameterized by $\textbf{w}$. The critic network consists of a linear feedforward layer, the encoder~\ref{sec:Encoder}, and two consecutive linear feedforward layers.


\section{Results}\label{sec:Results}
\subsection{TSP Validation}
\begin{figure}
    \centering
    \includegraphics[width=\columnwidth]{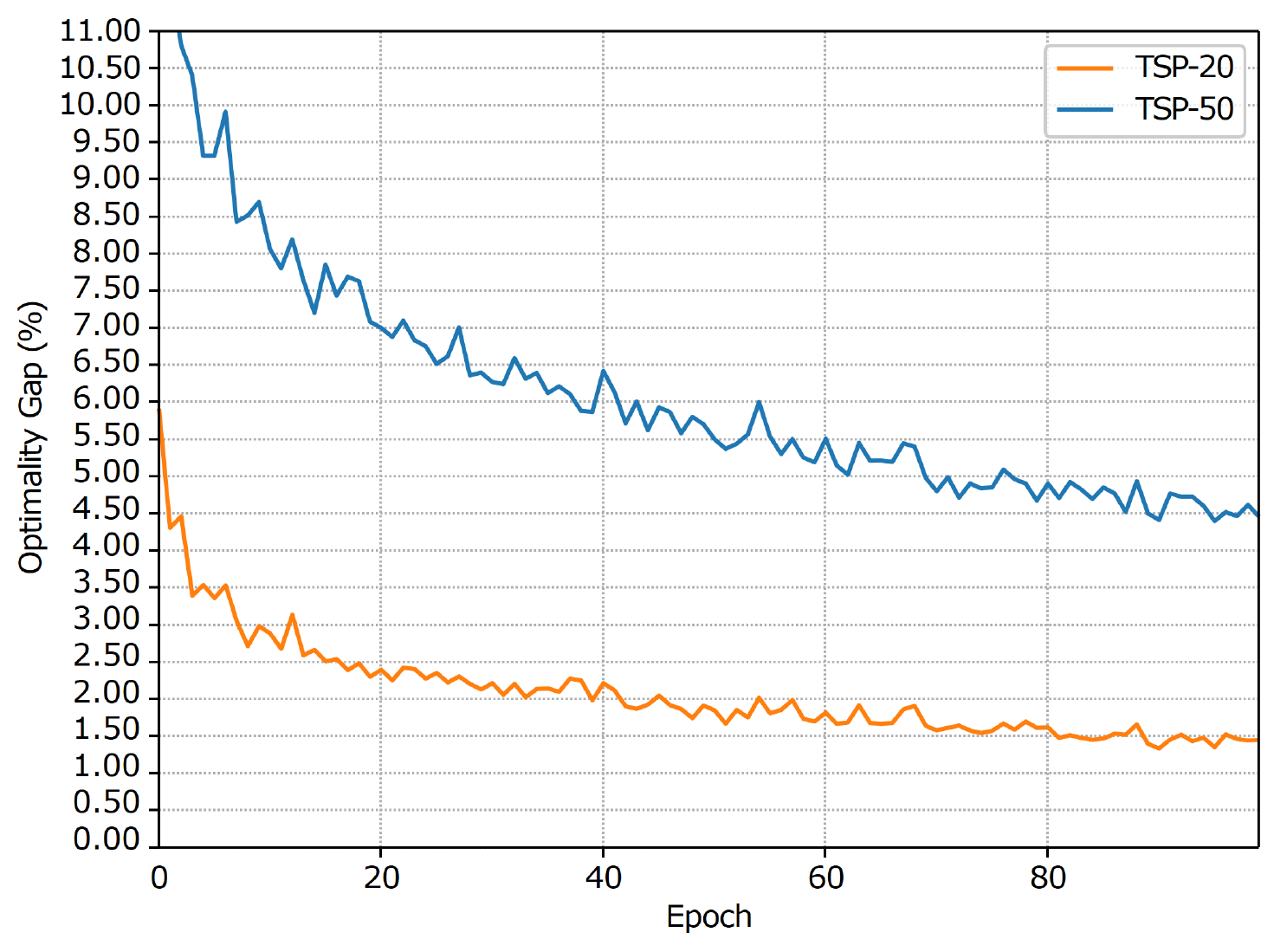}
    \caption{Pure TSP results.}
    \label{TSP_Results}
\end{figure}

The results from the first test, shown in Fig.~\ref{TSP_Results}, align with those reported by Joshi et al.~\cite{TSP_RL_2022} [Figure 11, TSP-50, L=3, d=128 (RL)]. The $100$ epochs in Fig.~\ref{TSP_Results} represent $12.8$ million samples in their study. This confirms that our environment can replicate the results of~\cite{TSP_RL_2022}. Although we used fewer samples, the goal was to validate that the framework, with the two MHA operations described in~\eqref{eq: h_s_C} and \eqref{eq:hK}, performs as expected.

\subsection{Decoder Functionality}
After confirming the system could solve the TSP, we conducted a second test. Here, starting and stopping points in each area were fixed, and the problem was treated as a pure TSP by the Concorde solver. The challenge here was in defining the edge matrix $E$, explained in Section~\ref{sec:ProblemFormulation}. Concorde can only provide solutions when the provided $E$ is symmetric with integer elements. However, $E$ in this test was not symmetric and the its elements were all float numbers due to map normalization, as described in Section~\ref{subsec: Map}. To address that, the elements of $E$ were scaled by 100 and rounded. Moreover, $E$ was symmetrized using the method described in \cite{symmetricisation}.
Note that it has been proven in~\cite{symmetricisation} that using the symmetrization method described preserves optimality. Specifically, the optimal solution to the newly obtained symmetric problem is also optimal for the original asymmetric problem. This guarantees that the Concorde solution would represent the true optimal solution and can therefore be treated as a ground truth.

\begin{figure}
    \centering
    \includegraphics[width=\columnwidth]{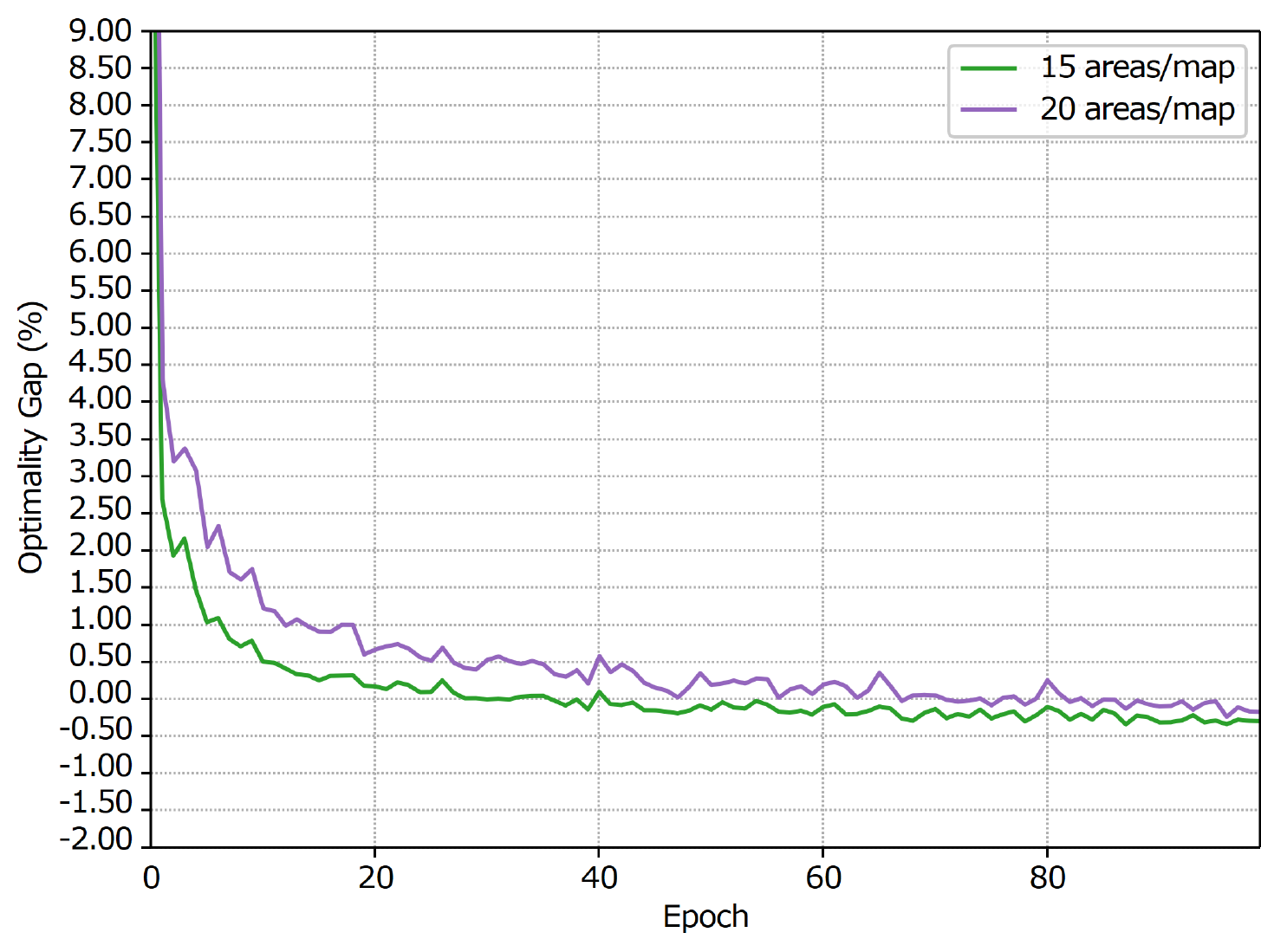}
    \caption{Optimality gap for the network choosing the pattern.}
    \label{TA_P1_Results}
\end{figure}

On the other hand, the same setup was applied in the RL environment, where the network was expected to learn the correct starting point, resulting in a $0\%$ optimality gap. The results, shown in \figurename~\ref{TA_P1_Results}, indicate that the agent successfully learned to solve the TSP under this constraint. Unlike the pure TSP test (\figurename~\ref{TA_P1_Results}), the agent achieved $0\%$ optimality gap in this test as having two decoders helped the agent better capture and comprehend the problem. Toward the end of the training, the model even produced slightly better results, reflected as a negative optimality gap in \figurename~\ref{TA_P1_Results}. This slight improvement is due to the approximations described earlier, which had to be done to obtain a solution from Concorde. The distances in the edge weight matrix (i.e., the distances from the exit point to the starting point) are
\begin{equation}
    \begin{aligned}
    a_{ij} = &\begin{cases} \| p_{i\_ exit} -p_{j_\_ start}\|_2, & \text { if } i \neq j \\ 0 & \text { otherwise. }\end{cases}
    \end{aligned} 
    \label{edge_weight_matrix}
\end{equation}

\subsection{System Assessment}
\begin{figure}
    \centering
    \includegraphics[width = \columnwidth]{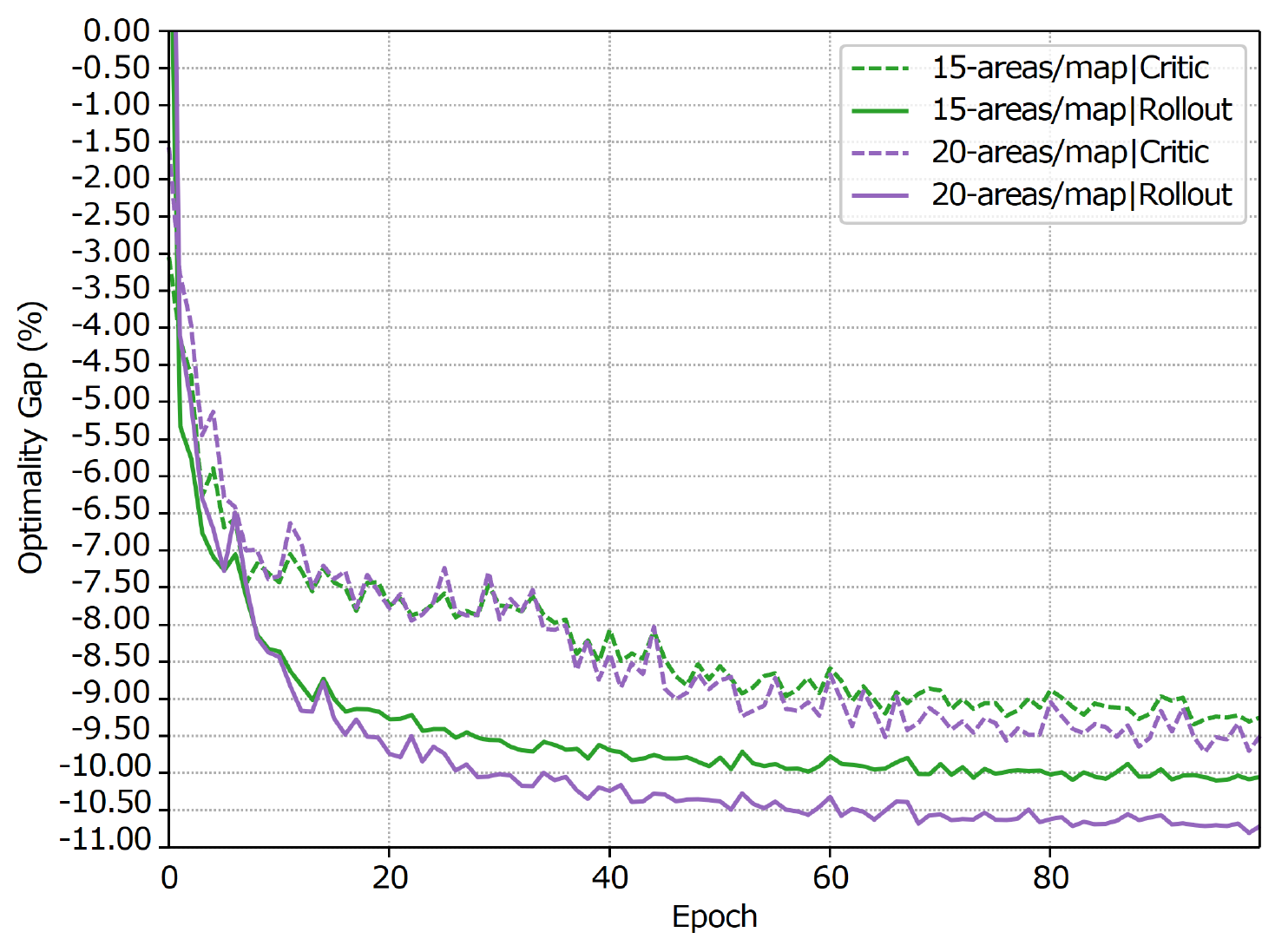}
    \caption{Optimality gap for the whole RouteFormer framework.}
    \label{Full_Results}
\end{figure}

After combining all decoder parts to test the full system, we found that fixing starting and stopping points yielded results matching the Concorde's solution, which can be considered the ground truth solution in \textit{this} case. As shown in \figurename~\ref{Full_Results}, the framework consistently achieved a negative optimality gap, as expected, with the gap increasing as the number of areas grew. This matches the expectation that the total distance becomes more sensitive to starting and stopping points with more areas. Additionally, testing with a critic baseline network showed that a rollout baseline provided better results, as learning with a fixed baseline is more effective than learning with a moving target.

To further assess the performance, we compared RouteFormer's solution against those obtained using well-known heuristic solvers like Google OR-Tools~\cite{ortools}, IBM CPLEX~\cite{IBM_CPLEX}, and  Lin-Kernighan-Helsgaun-3 (LKH-3)~\cite{Helsgaun_2017_LKH3}. These solvers can be used to solve the overall problem without any modification. \tablename~\ref{tab:multiplt_solvers_results} presents the comparative results and shows that RouteFormer achieved the best overall performance. Among the baseline solvers, LKH-3 yielded the best results and RouteFormer achieved an additional improvement of approximately $7\%$ over its solution. 

\begin{table*}
\centering
\caption{Costs from different solvers on evaluation datasets.}
\label{tab:multiplt_solvers_results}
\begin{tabular}{@{}ccccccccccc@{}}
\toprule
\multirow{2}{*}{\textbf{\begin{tabular}[c]{@{}c@{}}Number of\\ Areas\end{tabular}}} 
& \multicolumn{2}{c}{\textbf{Concorde}} 
& \multicolumn{2}{c}{\textbf{OR-Tools}} 
& \multicolumn{2}{c}{\textbf{CPLEX}} 
& \multicolumn{2}{c}{\textbf{LKH-3}} 
& \multicolumn{2}{c}{\textbf{RouteFormer}} \\ 

\cmidrule(lr){2-3} \cmidrule(lr){4-5} \cmidrule(lr){6-7} \cmidrule(lr){8-9} \cmidrule(l){10-11}

 & $\mu$ [m] & $\sigma$ [m] 
 & $\mu$ [m] & $\sigma$ [m] 
 & $\mu$ [m] & $\sigma$ [m] 
 & $\mu$ [m] & $\sigma$ [m] 
 & $\mu$ [m] & $\sigma$ [m] \\ \midrule

15 & 3.91 & 0.32 & 3.81 & 0.29 & 3.97 & 0.38 & 3.80 & 0.29 & \textbf{3.52} & 0.33 \\
20 & 4.40 & 0.33 & 4.25 & 0.30 & 5.09 & 0.56 & 4.21 & 0.28 & \textbf{3.92} & 0.34 \\ \bottomrule
\end{tabular}
\end{table*}

\subsection{Areas with Different Shapes}
To verify RouteFormer's ability to handle problems with areas of different shapes, we generated $1000$ samples with $20$ areas of random polygons. To simplify comparisons with other tools, the polygons were limited to $4$ random corners. Note that different number of corners could be used with the same architecture. 
For this test, CPLEX was not used as it takes over $10$ minutes to process one sample. 

\tablename~\ref{tab:different_shapes_results} summarizes the obtained results, whre both RouteFormer and LKH-3 achieved the best performance. RouteFormer matched the result of LKH-3, demonstrating its ability to handle areas of varying shapes while substantially reducing execution time. Whereas LKH-3 required approximately $240$ seconds to process all the samples, RouteFormer processed the same task in less than one second.

Unlike the results presented in \tablename~\ref{tab:multiplt_solvers_results}, RouteFormer did not outperform LKH-3 in terms of path cost for this specific test; instead, both solvers converged to the same solution. This is attributed to the pattern configurations defined in Section~\ref{sec:scenario}, which effectively reduced the feasible solution space and limited the potential for optimization gains. Crucially, however, RouteFormer reduced the execution time by $99.7\%$ compared to LKH-3. This efficiency is the deciding factor for real-world deployment, particularly because the execution relies entirely on the CPU. IoT nodes typically operate with limited onboard processing power and cannot afford the high number of CPU cycles required by iterative solvers like LKH-3. By drastically reducing the computational load, RouteFormer proves its suitability for these hardware-constrained environments, ensuring that efficient path planning does not come at the cost of the device's energy budget.

\begin{table*}
\centering
\caption{Costs and execution times for 20 areas of random shapes.}
\label{tab:different_shapes_results}
\begin{tabular}{@{}ccccccc@{}}
\toprule
\multirow{2}{*}{\textbf{\begin{tabular}[c]{@{}c@{}}Number of\\ Areas\end{tabular}}} 
& \multicolumn{2}{c}{\textbf{OR-Tools}} 
& \multicolumn{2}{c}{\textbf{LKH-3}} 
& \multicolumn{2}{c}{\textbf{RouteFormer}} \\ 

\cmidrule(lr){2-3} \cmidrule(lr){4-5} \cmidrule(l){6-7}

 & $\mu$ [m] & $\mu$\_time [s] 
 & $\mu$ [m] & $\mu$\_time [s] 
 & $\mu$ [m] & $\mu$\_time [s] \\ \midrule

20 & 3.95 & 5.009 & 3.91 & 0.24088 & \textbf{3.91} & 0.0006097 \\ \bottomrule
\end{tabular}
\end{table*}

\subsection{Scaling Capabilities}
To evaluate the model's capabilities with varying numbers of areas, we increased the number of areas within the unit square space to \(30\), \(40\), and \(50\). It is important to note that the model was trained using samples with maximum \(20\) areas. 

The results in Table~\ref{tab:results} demonstrate RouteFormer's superior ability to generalize to unseen problem instances. Despite being tested on larger problem sizes ($N=30,40$) than typically seen in training, RouteFormer consistently outperforms both the Concorde and LKH-3 solvers in terms of average path cost for scenarios with up to 40 areas. Specifically, at $N=40$ RouteFormer achieves a mean cost of $5.45$ m compared to $5.55$ m for LKH-3, proving its robustness in scaling to more complex environments. While LKH-3 marginally surpasses RouteFormer at $N=50$, $6.15$ m vs. $6.23$ m, respectively, this slight difference in solution quality must be viewed in the context of computational cost.

A comparison of execution times highlights a major operational difference. LKH-3 required between $0.24$ and $1.876$ seconds per sample to solve instances of $20$ to $50$ areas. On the other hand, RouteFormer completed the same tasks in $0.000577$ to $0.00297$ seconds per sample. This is an improvement of roughly three orders of magnitude. Given these results, RouteFormer offers a more practical solution for IoT scenarios, providing comparable path quality without the prohibitive computational costs associated with LKH-3.

\begin{figure}
    \centering
    \includegraphics[width=\columnwidth]{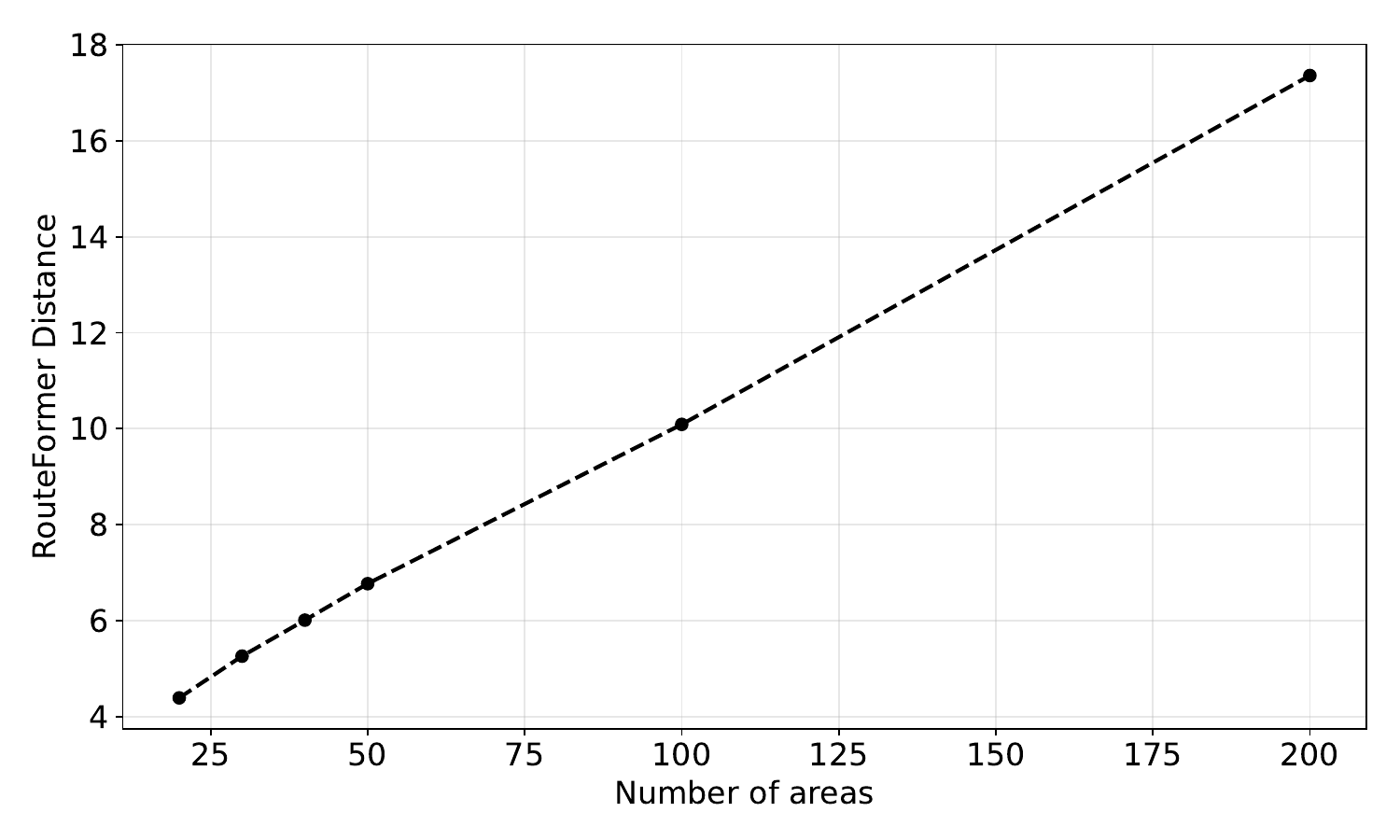}
    \caption{Scalability analysis of RouteForme.}
    \label{fig:scaling}
\end{figure}

To validate that this efficiency does not degrade at larger scales, we extended the evaluation beyond the initial test cases. \figurename~\ref{fig:scaling} illustrates the relationship between the number of areas and the average path cost generated by RouteFormer, scaling the problem size up to 200 areas. As observed, the model maintains a linear growth in path distance relative to the number of nodes. This linear trend is insightful, as it indicates that RouteFormer predicts consistent and stable trajectories even when handling instances $10$ times larger than those seen during training. The absence of exponential cost growth or erratic behavior confirms that RouteFormer generalizes effectively to large-scale problems, rather than being limited to the specific problem sizes encountered during training.

\begin{table}
\centering
\caption{Costs for a unit squared area.}
\label{tab:results}
\begin{tabular}{@{}ccccccc@{}}
\toprule
\multirow{2}{*}{\textbf{\begin{tabular}[c]{@{}c@{}}Number of\\ Areas\end{tabular}}} & \multicolumn{2}{c}{\textbf{Concorde}} &
\multicolumn{2}{c}{\textbf{LKH-3}} & \multicolumn{2}{c}{\textbf{RouteFormer}} \\ \cmidrule(l){2-7} 
 & $\mu$ {[}m{]} & $\sigma$ {[}m{]} & $\mu$ {[}m{]} & $\sigma$ {[}m{]} & $\mu$ {[}m{]} & $\sigma$ {[}m{]} \\ \midrule
20 & 4.39 & 0.34 & 4.20 & 0.26 & \textbf{3.90} & 0.33 \\
30 & 5.26 & 0.35 & 4.93 & 0.24 &\textbf{4.68} & 0.33 \\
40 & 6.01 & 0.39 & 5.55 & 0.23 &\textbf{5.45} & 0.33 \\
50 & 6.77 & 0.46 & \textbf{6.15} & 0.20 &6.23 & 0.34 \\ \bottomrule
\end{tabular}
\end{table}

Collectively, these results validate RouteFormer as a high-performance alternative to classical solvers like Concorde and LKH-3. The framework not only delivers competitive, and often superior, solutions with lower variance, but it also demonstrates stable generalization across problem sizes significantly larger than its training set. This consistency proves that the model has learned the underlying structural constraints of the routing problem rather than merely memorizing training patterns.

Beyond spatial optimization, a significant advantage of the proposed method is computational efficiency. By reducing inference time by nearly three orders of magnitude, RouteFormer effectively eliminates the processing bottleneck associated with onboard planning. This capability transforms path planning from a computationally expensive task into a near-instantaneous process, directly enabling energy-efficient autonomy for resource-constrained robotic agents.
\section{Conclusion}\label{sec:Conclusion}
In this work, we presented RouteFormer, a learning-based approach to single agent path planning that outperforms traditional baselines in both inference speed and adaptability. The results confirm that RouteFormer is not limited to the specific patterns seen during training; rather, it generalizes effectively to larger, unseen environments without the exponential cost growth associated with classical solvers. While it might match the solution quality of established methods like LKH-3 and Concorde, in few cases, its computational speed significantly outperform these methods in all the presented cases. Being nearly $600$ times faster than iterative solvers, RouteFormer is uniquely capable of operating within the tight processing margins of IoT hardware. This millisecond-level inference capability transforms path planning from a heavy computational burden into a negligible background task, directly translating to extended battery life and operational longevity. All this makes RouteFormer an ideal solution for resource-constrained IoT applications, where it enables fast, low-energy onboard planning without compromising the quality of the generated trajectories.

\begingroup
\bibliographystyle{IEEEtran} 
\bibliography{References}

\endgroup
\begin{IEEEbiography}[{\includegraphics[width=1in,height=1.25in,clip,keepaspectratio]{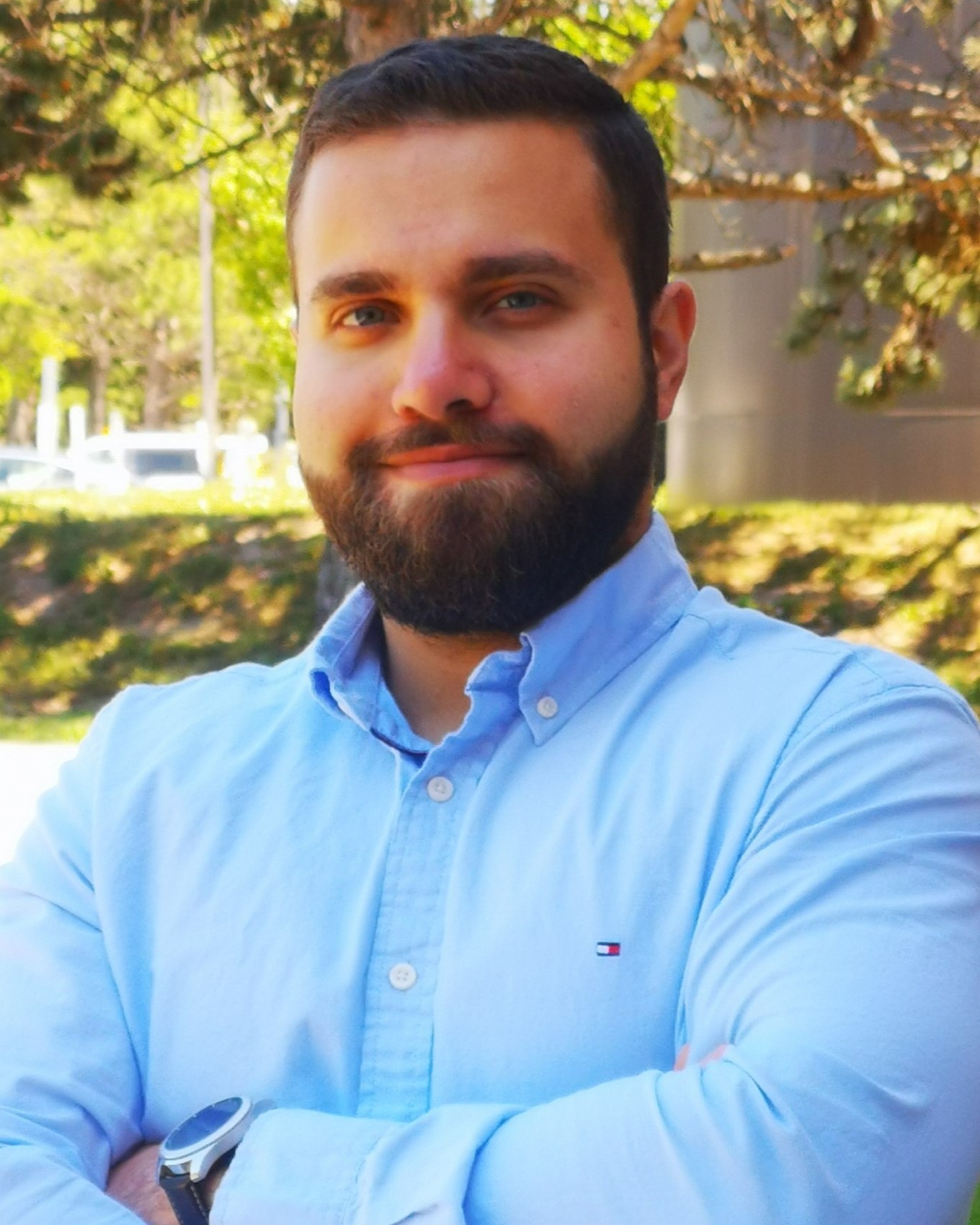}}]{Yazan Youssef} (GSM'23) received an M.Sc. degree in Electrical Engineering from the American University of Sharjah (AUS), Sharjah, UAE. He is now a Ph.D. candidate in the Electrical and Computer Engineering department at Queen’s University, Kingston, ON, Canada. His research interests include planning in autonomous systems and robotics, automation, and machine learning.
\end{IEEEbiography}

\begin{IEEEbiography}[{\includegraphics[width=1in,height=1.25in,clip,keepaspectratio]{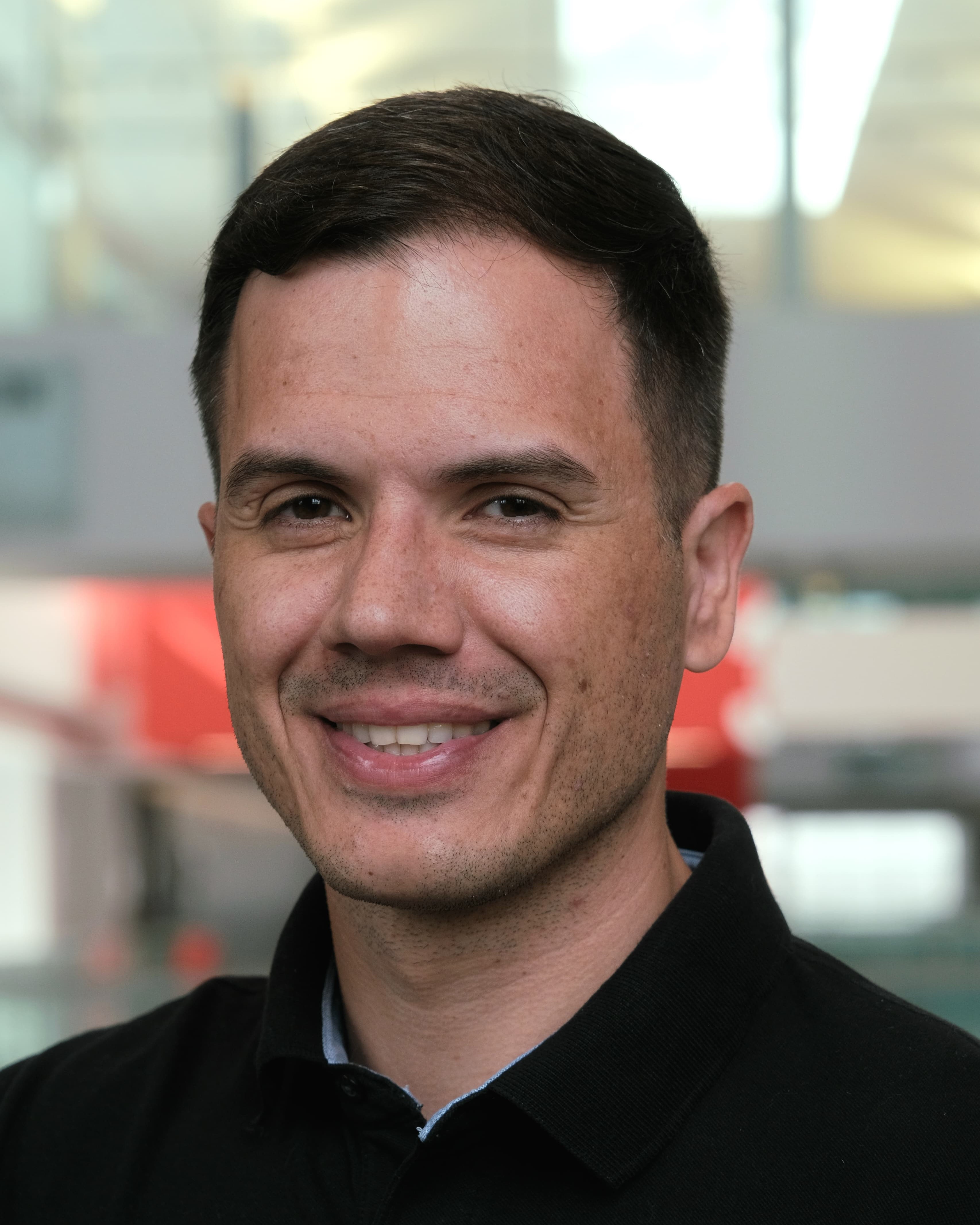}}]{Paulo Ricardo Marques de Araujo} (M'24) received a Ph.D. in Electrical and Computer Engineering from Queen’s University, Kingston, ON, Canada.
His broader research interests include autonomous systems, robotics, machine learning, and digital manufacturing.
\end{IEEEbiography}

\begin{IEEEbiography}[{\includegraphics[width=1in,height=1.25in,clip,keepaspectratio]{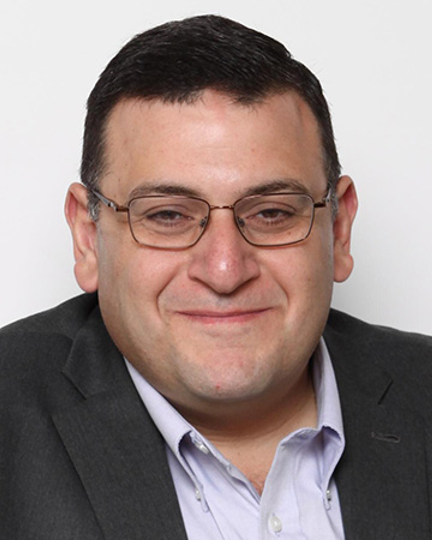}}]{Aboelmagd Noureldin} (SM'08) Dr. Aboelmagd Noureldin is a Professor and Canada Research Chair (CRC) at the Department of Electrical and Computer Engineering, Royal Military College of Canada (RMC), with Cross-Appointment at both the School of Computing and the Department of Electrical and Computer Engineering, Queen’s University. He is also the founding director of the Navigation and Instrumentation (NavINST) research group at RMC, a unique world-class research facility in GNSS, wireless positioning, inertial navigation, remote sensing and multisensory fusion for navigation and guidance. Dr. Noureldin holds a Ph.D. in Electrical and Computer Engineering (2002) from The University of Calgary, Alberta, Canada. In addition, he has a B.Sc. in Electrical Engineering (1993) and an M.Sc. degree in Engineering Physics (1997), both from Cairo University, Egypt. Dr. Noureldin is a Senior member of the IEEE and a professional member of the Institute of Navigation (ION). He published two books, four book chapters, and over 350 papers in academic journals, conferences, and workshop proceedings, for which he received several awards. Dr. Noureddin’s research led to 13 patents and several technologies licensed to industry in position, location and navigation systems.
\end{IEEEbiography}

\begin{IEEEbiography}[{\includegraphics[width=1in,height=1.25in,clip,keepaspectratio]{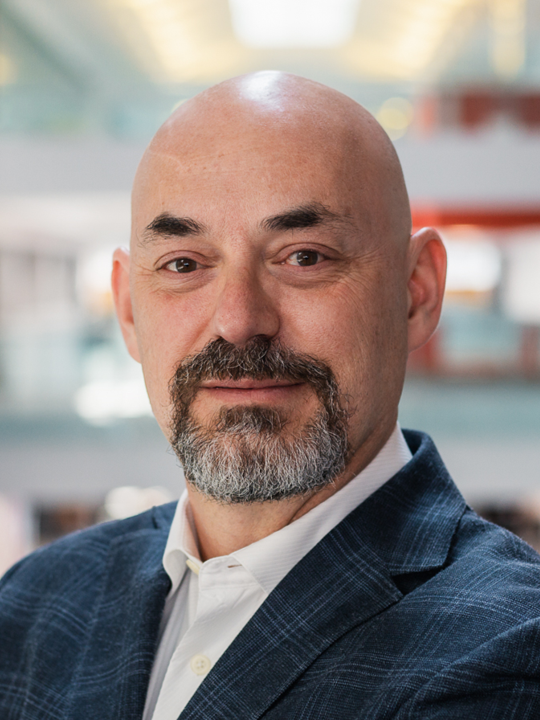}}]{Sidney Givigi} (SM'14) received a Ph.D. in Electrical and Computer Engineering from Carleton University, Ottawa, ON, Canada. He is now a Professor with the School of Computing of Queen's University, Kingston, ON. Sidney's research interests mainly focus on machine learning, autonomous systems, and robotics.
\end{IEEEbiography}

\end{document}